\documentclass[letterpaper, 10 pt, conference]{ieeeconf}  
\usepackage[utf8]{inputenc}
\usepackage{amsmath, amsfonts}
\usepackage{graphicx}
\usepackage{xcolor}
\usepackage{array}

\usepackage{wrapfig}
\usepackage{adjustbox}
\usepackage{xspace}
\usepackage{epstopdf}
\usepackage{xcolor}
\usepackage{float}
\usepackage{soul}
\usepackage{hyperref}
\usepackage{siunitx}
\usepackage{algorithm}
\usepackage{algcompatible}
\usepackage[capitalize]{cleveref}
\usepackage{booktabs}
\usepackage{multirow}
\usepackage[para]{threeparttable} 
\usepackage{array}
\usepackage{colortbl}
\usepackage{siunitx}
\usepackage{pifont}
\usepackage{booktabs}

\usepackage{eqexpl}
\eqexplSetDelim{=}

\usepackage{algpseudocode}
\usepackage{array}
\usepackage{tabularx}

\usepackage{wrapfig}
\usepackage[small]{caption}
\usepackage{subcaption}
\usepackage[noadjust]{cite}
\usepackage{comment}

\IEEEoverridecommandlockouts 

\newcommand{\systemname}{\textit{}\xspace}

\definecolor{darkgreen}{RGB}{0,100,0}

\title{\systemname Improving Sensing Coverage and Compliance of 3D-Printed Artificial Skins Through Multi-Modal Sensing and Soft Materials}

\author{Carson Kohlbrenner, Caleb Escobedo, Sayak Ray, Alexander Dickhans, Anna Soukhovei,\\Nickolaus Jackoski, Lyle Antieau, and Alessandro Roncone
\thanks{All authors are with the Human Interaction and Robotics [HIRO] Group, University of Colorado Boulder, 1111 Engineering Drive, Boulder, CO USA. This work is partially supported by NSF FW-
HTF-R grant \#2222952.
{\tt\small name.surname@colorado.edu}.}}
\begin{document}
\maketitle

\begin{abstract}

3D-printed artificial skins are a scalable approach to whole-body tactile and proximity coverage, but prior implementations have been limited to unimodal sensing and rigid materials. To improve the practical usability of 3D-printed artificial skins, we present a hybrid time-of-flight (ToF) and self-capacitance (SC) sensing skin that demonstrates multi-modal sensing integration, soft compliant coverings for impact absorption and pressure sensing, and a streamlined electrical interface between printed conductive traces and external electronics. We show that combining ToF and SC modalities enables contact detection, scene reconstruction, and pressure-correlated tactile responses with the compliant covering by deploying six artificial skin units with 40 sensing elements over an FR3 robot arm.
\end{abstract}


\section{Introduction}
 
Cameras situated around the workspace of a robot are the de facto standard for sensing the environment in unstructured operation, but this mode of sensing is impractical for detecting imminent contact on a highly dynamic system. The body of the robot itself often occludes cameras from seeing the moment of contact, leading to critical gaps in spatial awareness \cite{navarro2021proximity}. Artificial skins that can detect nearby objects from the exterior of the robot are a direct solution to combat occlusions, and the observable space they provide is determined by the sensing modality employed \cite{Dahiya2019LargeArea}.

Time-of-flight (ToF) and self-capacitance (SC) sensing are two commonly used modalities for measuring proximity \cite{scholz2024sensor}. ToF sensing artificial skins measure the time it takes a pulse of infrared light to reflect off a nearby surface to estimate distance and can have meter-scale sensing ranges \cite{kim2024armor}. Capacitive sensing skins on the other hand measure how nearby objects impact an electrostatic field to estimate proximity and are typically used for close range pre-touch sensing due to their wide but shallow coverage near the surface of the robot \cite{yang2024digital, Zhe2023self}. Both ToF and SC modalities have been integrated into modular sensor units that increase observability due to their complementary sensing coverages \cite{niquet2023prototype, tsuji2025wrap, yim2024electromagnetic}, yet scaling this multi-modal strategy to whole-body coverage requires a fabrication approach capable of conforming to the complex geometry of an entire robot body.

3D-printed multi-material artificial skins are a recent approach that use generative modeling to scale artificial skin units to large-area coverage and match the shape of a target robot’s links \cite{kohlbrenner2024gentact}. Prior 3D-printed skins have been demonstrated to cover large surface areas but have only been deployed with rigid components and single sensing modalities that do not offer continuous sensing coverage from close to far range \cite{kohlbrenner2026gentactprox, soukhovei2026form}. These limitations reduce their practical utility: unimodal sensors constrain the range and type of detectable events, and hard plastic coverings risk damage to robots and humans on contact. Additionally, ad hoc wiring between printed SC sensors and external electronics can cause signal drift when wires bend.

To advance 3D-printed artificial skin deployment, we demonstrate a multi-modal ToF and SC sensing artificial skin with large-area sensing coverage and soft compliant protection. This work advances the state-of-the-art for 3D-printed artificial skins by demonstrating: 1) multi-modal sensing via hybrid tactile and proximity sensing, 2) a flexible protective layer that houses internal wiring and electronics in a collision compliant manner, and 3) a threaded insert design that eliminates external wiring for printed SC sensors.

\begin{figure}
    \centering
    \includegraphics[width=0.95\linewidth]{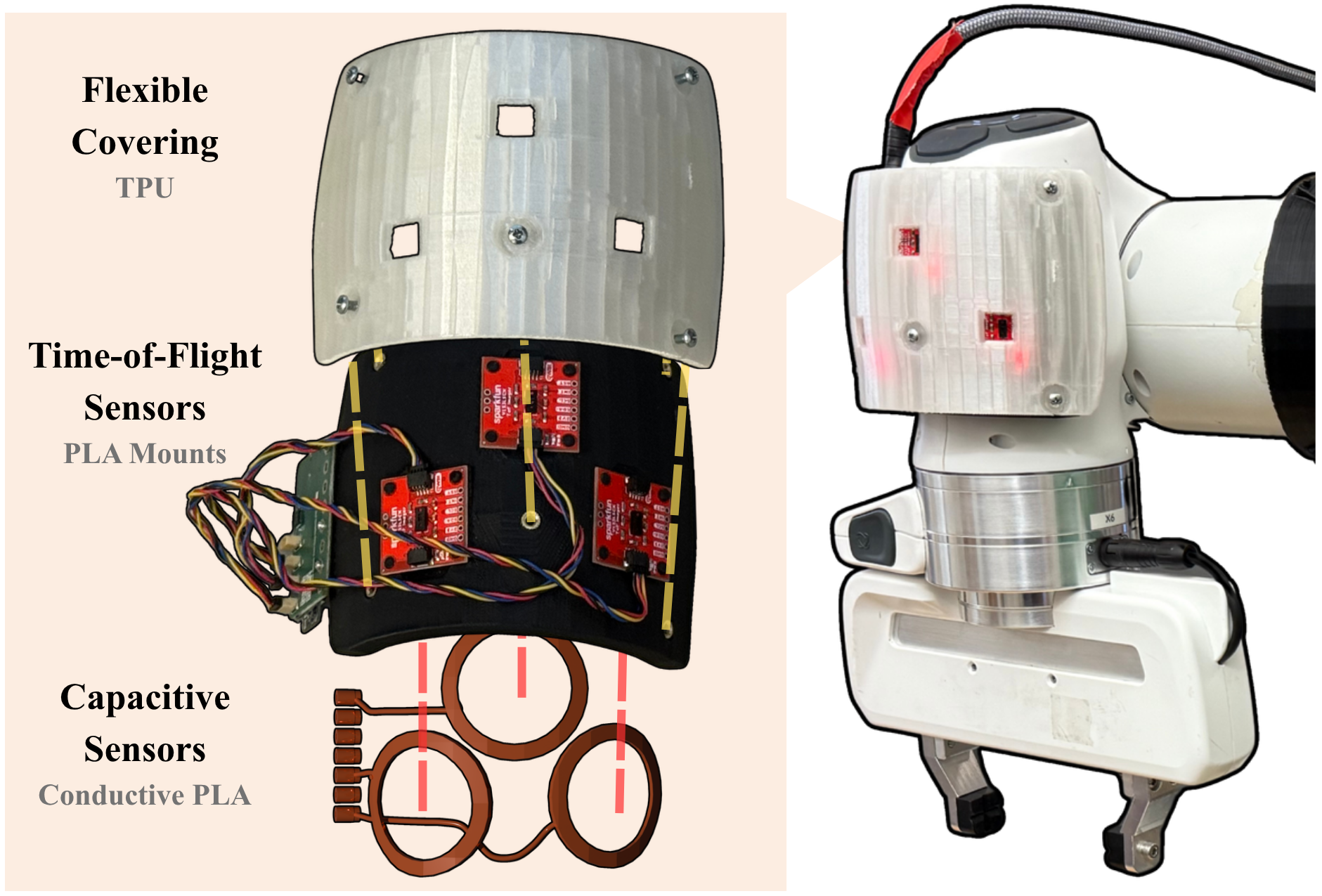}
    \caption{The hybrid sensors are composed of three layers: 1) SC sensors, 2) ToF sensor mounts, and 3) compliant soft covering.}
    \label{fig:cover}
    \vspace{-0.4cm}
\end{figure}
\begin{figure*}
    \centering
    \includegraphics[width=\linewidth]{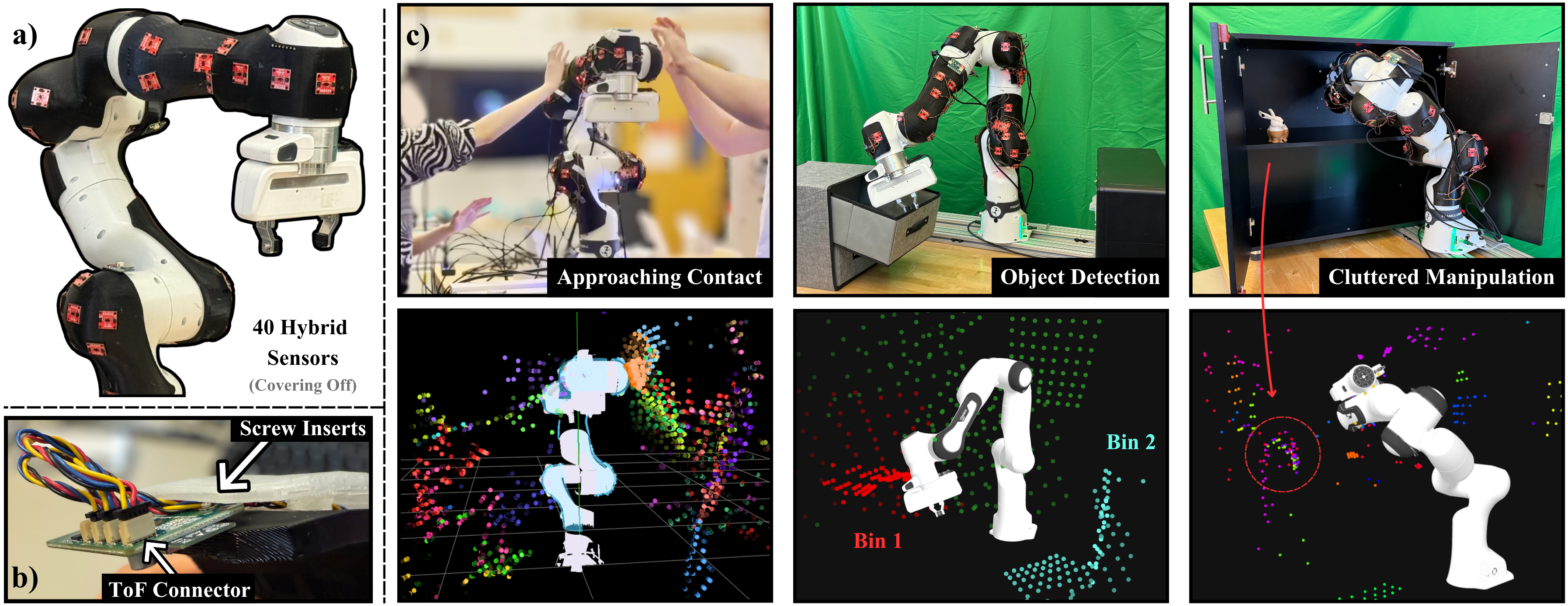}
    \caption{a) 40 hybrid sensing nodules distributed over the body of the FR3. b) Screw inserts were used for efficient wire management to the microcontroller. c) The depth data observed by the distributed sensors are demonstrated in various robotics scenarios.}
    \label{fig:results}
    \vspace{-0.2cm}
\end{figure*}

\section{Methods}
\subsection{Procedural Design}
The hybrid skin procedural design is implemented using Blender geometry nodes to automatically generate around a robot's CAD model and promote rapid design feedback. The procedural design expands on previous works that generate a form-fitting dermis around the robot's shape, then randomly distribute points along the surface of the dermis to instantiate capacitive sensor rings or custom PCB mounts at \cite{kohlbrenner2026gentactprox, soukhovei2026form}. We take advantage of the modular structure of the procedural design process to add the SC sensors and ToF mounts in union non-destructively. It should be noted that the ToF sensors interact with the electromagnetic field and add noise to the capacitive sensors. In an effort to reduce this effect, the capacitive sensors are hollowed out into rings around the ToF sensors to increase their coupling distance. 

The flexible covering is generated by extruding a copy of the dermis layer outwards by a user specified offset. Supports that connect the protective layer to the dermis layer are automatically distributed around the electronics as to not intersect with the mounts or capacitive sensors.

\subsection{Hardware}
Each skin unit was printed on a Prusa XL multi-nozzle fused deposition modeling (FDM) 3D printer using Protopasta conductive polylactic acid (PLA) filament for the wires and capacitive rings, standard PLA for the structural dermis, and flexible thermoplastic polyurethane (TPU) for the compliant covering. Our custom microcontroller uses an ESP32-C6 running at $\qty{160}{MHz}$ to measure capacitance at $42\pm 2$~Hz ($R\approx1M\Omega$) \cite{Fonseca2023AFP}. External wires to the SC sensors were eliminated by applying brass threaded inserts to the printed wire terminals that aligned and screwed in with the microcontroller pin-outs (\cref{fig:results}-b). The microcontroller featured seven SC pin-outs and four Qwiic connector ports daisy chained to Sparkfun VL53L5CX ToF imagers reporting 8x8 grids of proximity measurements up to $4$~m at $\qty{12}{Hz}$ \cite{vl53l5cx_datasheet}. The two modalities ran in parallel and were not fused.

\section{Results}
We demonstrate our approach by deploying six artificial skin units with a total of 40 individual hybrid sensing elements over the Franka FR3 robot arm (\cref{fig:results}-a). We reconstruct a dynamic scene into a point cloud using the skins' onboard proximity sensors (\cref{fig:results}-c).

\begin{table}
\centering
\begin{tabular}{l c c c}
\toprule
 & \textbf{No covering} 
 & \textbf{Covering (Rest)} 
 & \textbf{Covering (Squeeze)} \\
\midrule
\textbf{With ToF}    
& $50 \pm 20$
& $13 \pm 3$ 
& $22 \pm 5$ \\
\vspace{-0.1cm}\\
\textbf{No ToF} 
& \shortstack{$120 \pm 20$} 
& \shortstack{$37 \pm 4$} 
& \shortstack{$45 \pm 3$} \\
\bottomrule
\end{tabular}

\caption{Average contact SNR of the capacitive sensors in \cref{fig:cover}.}
\label{tab:snr_results}
\vspace{-0.5cm}
\end{table}

\subsection{Multi-Modal Interference}
We collected capacitive sensor readings from the sensor shown in \cref{fig:cover} with and without ToF sensors and the flexible covering to evaluate signal quality and interference. For each permutation, the signal to noise ratio (SNR) was estimated using $\text{SNR} = |\mu_n - \mu|/\sigma_n$ with the average inactive signal $\mu_n$ when no objects are within $\qty{1}{m}$, the inactive signal standard deviation $\sigma_n$, and the average active signal $\mu$ when a human palm is resting on the sensor.

Each data point is the average from all three SC rings over 1000 samples. The results of this experiment are shown in \cref{tab:snr_results}, which reveals that the SC sensors can detect contact above the minimum SNR threshold of 7 in all configurations \cite{2010AN1T}. When the flexible covering was attached, the mean active signal was reduced in magnitude at rest due to the hand being positioned further away from the SC sensors than without the covering. The active signal was observed to increase with pressure as it was squeezed, demonstrating that the sensor's tactile response is correlated with contact detection and pressure.

\section{Conclusion}
In this work, we introduce a 3D-printed artificial skin that uses hybrid SC and ToF sensing modalities to increase observability and soft printed materials to increase compliance on collision and pressure sensitivity. The procedural backbone of the sensor design allowed the modalities to be combined with ease, and we demonstrated that combining the sensing modalities does not prevent the SC sensors from reliably detecting contact and pressure. Limitations of this work include SC sensors only reliably detecting conductive objects and the lack of signal alignment methods between SC and ToF modalities. Future steps for this work include formally quantifying the hand-off zones between the SC and ToF sensors and formulating methods for fusing the ToF and SC signals in heterogeneous distributions. 

\bibliographystyle{IEEEtran}
\bibliography{references}
\end{document}